\documentclass[runningheads]{llncs}
\usepackage{graphicx}

\usepackage{amssymb}
\usepackage{amsmath}
\usepackage{soul}
\usepackage{times}
\usepackage{empheq}
\usepackage{latexsym}
\usepackage{booktabs}
\usepackage{multirow}
\usepackage{subfigure}

\usepackage{color}

\begin{document}
\title{Dependency Structure for News Document Summarization
% \thanks{Supported by organization x.}
}
% \author{Anonymous}
% \institute{Anonymous}
\author{Congbo Ma\inst{1} \and
Wei Emma Zhang\inst{1} \and
Hu Wang\inst{1} \and
Shubham Gupta\inst{1} \and
Mingyu Guo\inst{1}}
\authorrunning{C. Ma et al.}
\institute{The University of Adelaide, Adelaide, Australia\\
\email{\{congbo.ma, wei.e.zhang, hu.wang, shubham.gupta, mingyu.guo\}@adelaide.edu.au}}

\maketitle              % typeset the header of the contribution
\begin{abstract}
In this work, we develop a neural network based model which leverages dependency parsing to capture cross-positional dependencies and grammatical structures. With the help of linguistic signals, sentence-level relations can be correctly captured, thus improving news documents summarization performance. Empirical studies demonstrate that this simple but effective method outperforms existing works on the benchmark dataset. Extensive analyses examine different settings and configurations of the proposed model which provide a good reference to the community.

\end{abstract}
\section{Introduction}

Multi-document summarization (MDS) is a critical task in natural language processing (NLP) aiming at generating a brief summary from a set of content-related documents. MDS enjoys a wide range of real-world applications, including the summarization of news, scientific publications, emails, product reviews, medical documents and software project activities \cite{ma2020multi}. There are two main types of summary generation: extractive summarization and abstractive summarization. Extractive summarization selects salient sentences from the original texts to form the summaries. In abstractive summarization, machine generates summaries from its understanding of the contents and it is more similar to human-written summaries. This characteristic makes abstractive summarization a challenging task. With the development of deep learning techniques, neural network based models are widely applied in abstractive multi-document summarization \cite{alexander2019multi,liu2019hierarchical,wei2020leveraging}. These techniques enable better performances of the MDS models and significantly prosper the research of Multi-document summarization. 
The neural network based models have strong fitting capabilities. One recent architecture Transformer \cite{ashish2017attention} shows strong performances in various natural language processing tasks and is also adopted in multi-document summarization \cite{alexander2019multi}. Transformer has natural advantages for parallelization and could retain long-range relations among tokens. However, similar to other neural network models, existing Transformer-based MDS models neglect linguistic knowledge in the sentences that may affect the summary qualities greatly.

In the family of linguistic information, dependency parsing is one of the most important knowledge. It retains the syntactic relations between words to form a dependency tree and offers discriminate syntactic paths on arbitrary sentences for information propagation through the tree \cite{kai2019aspect}.  %
Inspired by the success of dependency parsing applied in variety NLP tasks,
we introduce a generic framework to combine dependency parsing with the Transformer architecture to better capture the dependency relations and grammatical structures for abstractive summary generation. Our work is one of the first few to simultaneously leverage linguistic information and Transformer-based models for multi-document summarization. 
More specifically, the dependency information is processed to linguistic-guided attention. Later on, it is further merged with multi-head attention for better feature representation. With the assist of the linguistic signals, sentence-level relations can be correctly captured.
The experimental results on the benchmark dataset indicate the proposed approach brings improvements over several strong baselines. The contributions of this work are summarized as follows:

\begin{itemize}
\item 
We propose a simple yet effective linguistic-guided attention mechanism to incorporate dependency relations into Transformer's multi-head attention. The proposed linguistic-guided attention can be deployed seamlessly to multiple mainstream Transformer-based MDS models to improve their performances.
\item We evaluate and compare the proposed model with several baseline models and multiple competitive MDS models. The results demonstrate that the models equipped with the linguistic-guided attention receive superior performances over the comparing models.
\item We provide extensive analysis on various settings and configurations of the two versions of the proposed model ParsingSum. These results could help understand the intuitiveness of the proposed model and could serve as an informative reference to the multi-document summarization research community. 

\end{itemize}

\section{Related Work}
Abstractive multi-document summarization task gains increasing attention in recent years. Compared to single document summarization, multi-document summarization requires the model to generate summaries from multiple documents in a more comprehensive manner. 
However, abstractive summarization have met fewer successes due to the lack of sufficient datasets, non-trivial effort to extend sequence-to-sequence structure from single document summarization to multi-document summarization \cite{liu2019hierarchical,wei2020leveraging} and lack of approach to address cross-document relations.  Liu et al. \cite{peter2018generating} constructed a large-scale multi-document summarization dataset and adopted Transformer model to multi-document summarization. The selected top-K tokens are fed into a decoder-only sequence transduction to generate the Wikipedia articles. Based on this work, Yang et al.\cite{liu2019hierarchical} proposed a hierarchical Transformer (HT) architecture that contains token-level and paragraph-level Transformer layers to capture cross-document relationships. Inspired by HT, Li et al. \cite{wei2020leveraging} incorporated graph representation into encoding and decoding layers to capture rich relations among documents and lead the summary generation process. Jin et al \cite{han2020multi} proposed a Transformer-based multi-granularity interaction network to unify the extractive and abstractive multi-document summarization. However, these Transformer-based models do not take consider dependency relations in the source documents into account.

One research direction for text summarization is utilizing extra knowledge to assist summary generation process. Daniel et al. \cite{leite2007extractive} suggested that the linguistic knowledge could help improve the informativeness of document summary. Sho et al.\cite{takase2016neural} proposed an attention-based encoder-decoder model that adopts abstract meaning representation parser to capture structural syntactic and semantic information. The authors also pointed out that for natural language generation tasks in general, semantic information obtained from external parsers could help improve the performance of encoder-decoder based neural network model. Patrick et al. \cite{patrick2019structure} adopted named entities and entity coreferences for summarization problem. Jin et al. \cite{jin2020semsum} enriched a graph encoder with semantic dependency graph to produce semantic-rich sentence representation. Song et al. \cite{song2020joint} presented a LSTM-based model to generate sentences and the parse trees simultaneously by combining a sequential and a tree-based decoder for abstractive summarization generation. Different from previous work, in this paper, we propose a generic framework to combine the linguistic guidance with powerful architecture for summarization.

\section{Linguistic-guided Multi-document Summarization} \label{sec: Model Description}

In this work, we propose a linguistic-guided approach that leverage dependency parsing for abstractive multi-document summarization. We first overview the proposed model in Section \ref{sec:overview}. Then we focus on the detailed steps to form dependency information matrices in Section \ref{Dependency Information Matrix} and how to incorporate this linguistic information into the attention mechanism in Section  \ref{Linguistic Guided Encoding Layer}.  

\subsection{Model Overview}
\label{sec:overview}

\textbf{Problem Definition}. Given a set of documents $D=\left ( d_{1},  d_{2},..., d_{p} \right )$, where $p$ is the number of documents, multi-document summarization is to generate the correct summary $Sum$ distilling knowledge from the document set. In our linguistic-guided model, we further define that for any sentence $s_{ij}$ of a document $d_{j}$, an external dependency parser generate a tree $T_{ij}$. Our task can be defined as $D+T \rightarrow  Sum$ that is to find a way to combine the knowledge delivered by the tree for better summary generation.

\begin{figure}[t]
\centering
\includegraphics[width=0.48\textwidth]{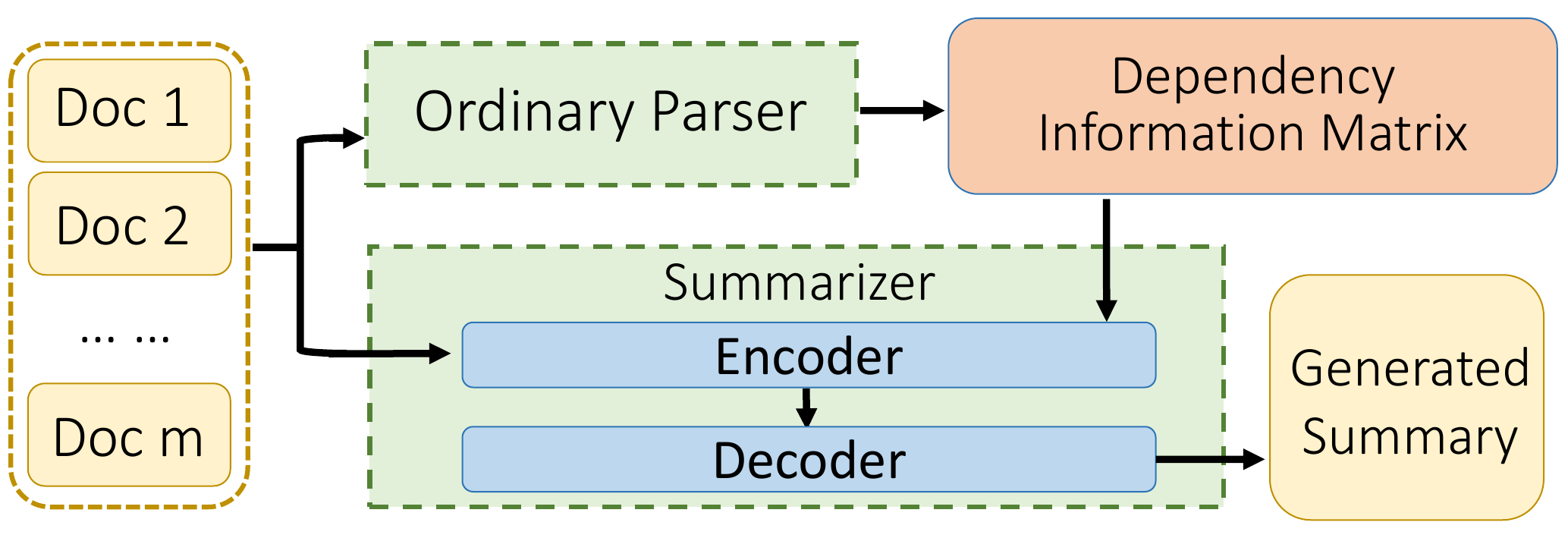}
\caption{The model framework of ParsingSum. The set of documents are first fed into the encoder to generate the representation. In  the meantime, these documents are input to a dependency parser to produce their sentence dependency information. The dependency information matrix will be further processed into a   linguistic-guided attention mechanism and then fused with Transformer's multi-head attention to guide the downstream summary generation.}
\label{fig:parsingsum_sys}
\end{figure}

Figure \ref{fig:parsingsum_sys} presents the framework of our proposed model, named ParsingSum. The model follows an encoder-decoder architecture \cite{bahdanau2015neural}. Our proposed model is generic and flexible to be hanged in different Transformer structures. 
Inside the model, the encoder is a representation learner to learn distinctive feature representations from the source documents and decoder is able to decipher representations into language domain for summary generation. More concretely, the document sets are first fed into the an Transformer-based encoder for feature representation. At the same time, the documents are passed into an external dependency parser to fetch the dependency relations. In the encoder, these relations will be processed into a linguistic-guided attention mechanism to further fuse with multi-head attention later on. With the assist of the linguistic information, the model is able to  grasp the linguistic relations of the input documents to guide the summary generation. In this paper, we focus on how to integrate parsing information into the model for better representation learning in the encoder. Followed by ashish et al. \cite{ashish2017attention}, we build a similar decoder.

\subsection{Dependency Information Matrix} \label{Dependency Information Matrix}
Dependency parsing is a family of grammar formalisms playing an important role in contemporary speech and language processing. It is appropriate to adopt it to deal with languages that are morphological-rich with unrestrained ordering of words. Given a sentence, dependency parsing extracts a dependency tree that represents its grammatical structure and defines the relations between \textit{head} words and \textit{dependent} words. This information could be utilized to guide the summarization.

We generate the dependency tree using an existing dependency parser \cite{dozat2017deep} for each input sentence and obtain a set of dependency trees. These trees contain dependency relations between any pair of dependent words in the given sentences. We record the trees into a matrix by $T_{ij}$. Let $\mathbb{P}$ denotes the dependency information matrix for the given input sentences $s_{ij}$, where $\mathbb{P}_{ij}$ indicates the dependency weights between word $i$ and word $j$. We simplify the weights definition as shown in Eq.(1).

\begin{equation} \small
\mathbb{P}_{ij} = 
\begin{cases}
1& \text{dependency relation exists}\\
0& \text{no dependency relation}
\end{cases}
\end{equation}

In the process of constructing dependency information matrix, our proposed method ignores the direction of \textit{head} word and \textit{dependent} word. For other pairs, as long as there is a dependency relation between two words, the dependency information matrix is assigned a value of 1, otherwise it will be set to 0. We hope to keep all dependency relations between the pair words in a simple yet effective manner.

\begin{figure*}[t]
\centering
\includegraphics[width=1.0\textwidth]{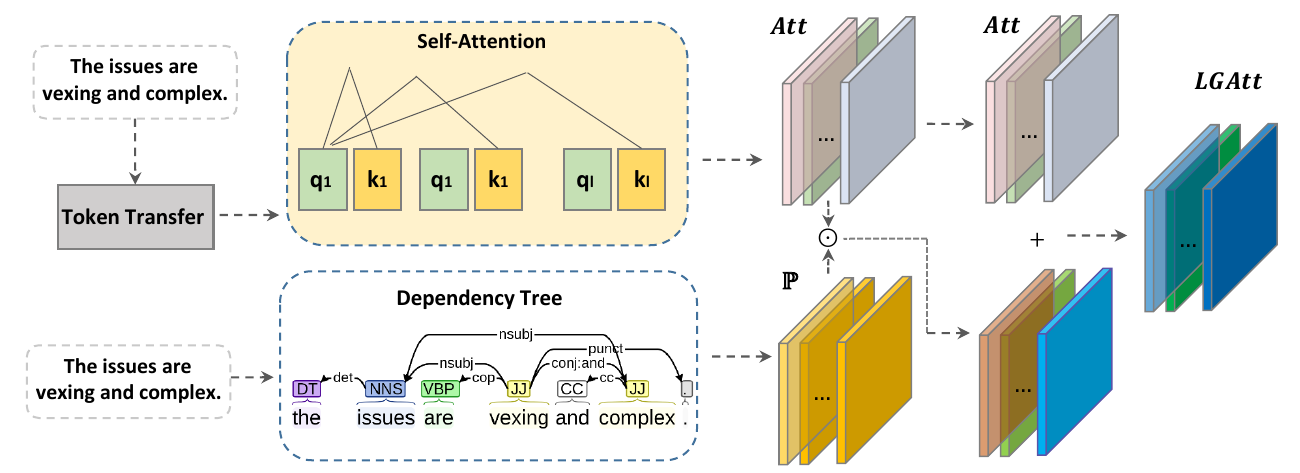}
\caption{The Linguistic-Guided Attention Mechanism. The given exemplary sentence \textit{The issues are vexing and complex.} is from Multi-News dataset \cite{alexander2019multi}. Different properties of vocabularies and relations between words are included in the parsing information. At the same time of generating self-attention, the sentence is feed into an external parser to fetch the dependency tree. Later on, the dependency tree are processed into a dependency information matrix $\mathbb{P}$. Then, it is merged into the original multi-head attention.}
\label{fig:parsing_example}
\end{figure*}

\subsection{Linguistic-Guided Encoding Layer} \label{Linguistic Guided Encoding Layer}
In order to process source documents effectively and preserve salient source relations in the summaries, we propose a novel linguistic-guided attention (LGA) mechanism that can extend the Transformer architecture \cite{ashish2017attention} in the encoding process.
Figure \ref{fig:parsing_example} depicts this mechanism on a exemplary sentence from  Multi-News \cite{alexander2019multi} dataset. 
LGA can joint the dependency parsing information and the source documents to generate semantic rich features in a complementary way. The linguistic-guided attention mechanism can be viewed as learning a graph representation for the sentences from the input documents. We model the dependencies across source documents with linguistic-guided attention. For the input token $x_{i}\in s(i=1,2,...,l))$ and attention head $head_j \in Head (j = 1,2,...,z)$. We have:

\begin{equation} \small
\begin{aligned}
q_{i, head_j}=W^{q, head_j} \cdot x_{i} \\
k_{i, head_j}=W^{k, head_j} \cdot x_{i} \\
v_{i, head_j}=W^{v, head_j} \cdot x_{i}
\end{aligned}
\end{equation}

\noindent where $W^{q, head_j}$, $W^{k, head_j}$, $W^{v, head_j}$ $ \in \mathbb{R}^{n*m}$ are weight matrices. $q_{i, head_j}$, $k_{i, head_j}$, $v_{i, head_j}$ $\in \mathbb{R}^{n*i}$ are sub-query, sub-key and sub-values in different head. We than concatenate these sub-query, sub-key and sub-values respectively.

\begin{equation} \small
\begin{aligned}
q_{i}=concat(q_{i, head_1}, q_{i, head_2}, ..., q_{i, head_z}) \\
k_{i}=concat(k_{i, head_1}, q_{i, head_2}, ..., q_{i, head_z}) \\
v_{i}=concat(v_{i, head_1}, q_{i, head_2}, ..., q_{i, head_z})
\end{aligned}
\end{equation}

\noindent where $q_{i}, k_{i}, v_{i} \in {R}^{z*i} $ are corresponding key, query and value for later attention calculation. The linguistic-guided attention merge the semantic dependency information $\mathbb{P}$ with multi-head attention. 

\begin{equation} \small
LGAtt_{ij}=\left(\alpha \mathbb{P}_{i j}+\mathbb{I}\right) \odot att_{i j}
\end{equation}

\begin{equation} \small
att{_{ij}}=softmax\left ( \frac{q_i \cdot k_j}{\sqrt{d_{head}}} \right )
\end{equation}

\begin{equation} \small
Context_{i}=\sum_{j} LGAtt_{ij} \cdot v_{j}
\end{equation}

\noindent where  $\alpha$ is a trade-off hyper-parameter to balance the linguistic-guide information and multi-head attention map, $\mathbb{I} \in {R}^{z*i} $ is an identity matrix, $d_{head}$ is the dimension of heads, $\odot$ is element-wise Hadamard product, $Context_{i} \in {R}^{z*n} $ representation the context vectors generated by linguistic-guide attention. Later on, two layer-normalization operations are applied to $Context_{i}$ :

\begin{equation} \small
 \hat{x}{_{i}}=LayerNorm\left ( k_{i}+FFN\left (k_{i} \right ) \right )
\end{equation}

\begin{equation} \small
k_{i}=LayerNorm\left ( x_{i} + Context_{i}  \right )
\end{equation}

\noindent $\hat{x}{_{i}}$ denotes the output of next encoding layers, where FFN is a two-layer feed-forward network with ReLU as activation functions. Then, the learner feature representations are passed to the decoder layers for summary generation. As the components of the decoder are similar to Flat Transformer structure \cite{ashish2017attention}, we focus on the construction of linguistic-guided attention in encoding layers.

\section{Experiments}

In this section, we report the effectiveness of the proposed Linguistic-guided attention. We first introduce the experimental settings and then overall model performances are presented. We further provide extensive analysis on how to select a suitable fusion weights in linguistic-guided attention, as well as the influence of batch size. Later on, discussion on different fusion methods and its visualization are conducted.

\subsection{Datasets}

Multi-News \cite{alexander2019multi} is a large-scale dataset containing various topics in news articles for multi-document summarization. It includes 56,216 article-summary pairs and it is further scattered with the ratio 8:1:1 for training, validation and test respectively. Each document set contains 2 to 10 articles with a total length of 2103.49 words. The golden summaries length is 263.66 on average.

\subsection{Evaluation Metric}
We evaluate the proposed model and compared its performances with multiple state-of-the-art models using ROUGE scores \cite{lin2004rouge}. Unigram and bigram overlap (ROUGE-1 and ROUGE-2 scores) are adopted to indicate the literal quality of generated summaries. Besides, the longest common subsequence measuring of ROUGE (ROUGE-L) enables us to measure the similarity of the two text sequences in the sentence-level and the fluency of generated summaries. ROUGE F1 scores (denoted as ROUGE-F) are considered in our work.

\subsection{Models for Comparison}
We compared ParsingSum with the following models: LexRank computes the importance of a sentence based on the concept of eigenvector centrality in a sentence graph  \cite{gunes2004lexrank}. TextRank is a graph-based ranking model \cite{rada2004textrank}. Maximal Marginal Relevance (MMR) algorithm \cite{Jaime1998the} considers the importance and redundancy of a sentence in a complementary way to decide whether to add the sentence to the summary. BRNN is an bidirectional RNN based model for summarization task. Flat Transformer (FT) is a Transformer-based encoder-decoder model on a flat sequence. Hi-MAP \cite{alexander2019multi} is an end-to-end model that incorporates MMR into a pointer-generator network. Hierarchical Transformer (HT) \cite{liu2019hierarchical} is an abstractive summarizer that can capture cross-document relationships via hierarchical Transformer encoder and flat Transformer decoder.

\subsection{Experimental Settings and Hyper-parameters}

We equipped the proposed linguistic-guided attention model with both Hierarchical Transformer (HT) and Flat Transformer (FT) leading to two versions of the proposed model: ParsingSum-HT and ParsingSum-FT$\footnote {The default training steps for FT is 50,000 \cite{alexander2019multi}. We only run 20,000 step and achieve promising results on FT and ParsingSum-FT. The performance would be better if continue training.}$. 
For ParsingSum-HT, we followed the implementation of the Hierarchical Transformer model by using six local Transformer layers and two global Transformer layers  with eight heads. For ParsingSum-FT, we followed Flat Transformer model settings and adopt four encoder layers and four decoder layers. 
Before training, we pre-processed the data by lowercase all English characters, and surround each target sentence with a delimiter.  
We also trunk 400 tokens of target documents and trunk 24 blocks of source documents in order to fit the models. 
For training, we set batch sizes as 13,000 and use \textit{Adam} optimizer ($\beta1$=0.9 and $\beta2$=0.998). 
The dropout rates of both encoder and decoder are set to 0.1. 
The trade-off hyper-parameter $\alpha$ is 1. 
The initial learning rate is set to $1 \times 10^{-3}$. The first 8000 steps are trained for warming up and the models are trained with a multi-step learning rate reduction strategy. 
% In our experiments, the model accumulates gradients and updates at once every four iterations. Maximum 250 tokens and minimum 20 tokens should be generated. 

\subsection{Overall Performance}

\begin{table} \small
\centering
\caption{Models Comparison in ROUGE-F. For each part, ROUGE-1, ROUGE-2 and ROUGE-L scores are examined. The best result for each column are in bold.}
\label{tab:model_comparison}
\setlength{\tabcolsep}{3mm}{
\begin{tabular}{c|ccc}
\hline
\multirow{2}{*}{Models}      & \multicolumn{3}{c}{ROUGE-F}  \\
            & 1       & 2       & L         \\ \hline
LexRank     & 37.92   & 13.10    & 16.86    \\
TextRank    & 39.02   & 14.54   & 18.33     \\
MMR         & 33.20    & 9.89    & 16.41   \\
BRNN        & 38.36   & 13.55   & 19.33      \\
FT          & 42.98   & 14.48   & 20.06    \\
Hi-MAP      & 42.98   & 14.85   & 20.36   \\
HT    & 36.89   & 12.76   & 20.37    \\ \hline
ParsingSum-HT (Ours)  & 37.34  & 13.00  & 20.42      \\
ParsingSum-FT (Ours)  & \textbf{44.32}  & \textbf{15.35}   & \textbf{20.72} \\ \hline
\end{tabular}}
\end{table}

We evaluated the proposed model ParsingSum with two versions and compared its performances with different MDS models.  For fair comparision, we reran all the comparing models on the same device with the same configuration settings\footnote{In this case, our results may be different with what the models' papers reported.}
As shown in the table, we perceived the ParsingSum-HT model receives higher ROUGE scores (across all ROUGE-1, ROUGE-2 and ROUGE-L scores) steadily compared to the original HT model. The Linguistic-Guided Attention helps the model raise 0.45 on ROUGE-1 score and 0.24 on ROUGE-2 score, respectively. Similar phenomenon shows on the  ParsingSum-FT model. More specifically, ParsingSum-FT surpasses FT model 1.34 on ROUGE-1 score, 0.87 on ROUGE-2 score and 0.66 on ROUGE-L score.

It is worth to note that the proposed ParsingSum-FT is able to outperform its baseline (i.e., FT model) by a large margin and also  receives the highest ROUGE-L score across all the comparing methods. It indicates the outstanding capability of ParsingSum models to retain the intention of original documents when generating summaries.
We further analyzed the effects of the trade-off parameter $\alpha$ and batch size in ParsingSum. We also examined and discussed different ways to incorporate parsing information into the ParingSum model.

\begin{table}[t] \small
\centering
\caption{The Analysis of Fusion Weights of Linguistic-guided attention on HT and FT based model. The results are shown in ROUGE-1, ROUGE-2 and ROUGE-L scores of ROUGE-F. The best result for each column within a section are bolded.}
\label{tab:parsing_weight_HT_b13000}
\setlength{\tabcolsep}{2.5mm}
\begin{tabular}{c|ccc}
\hline
\multirow{2}{*}{Models} & \multicolumn{3}{|c}{ROUGE-F} \\ 
                        & 1       & 2       & L         \\ \hline
HT                      & 36.09   & 12.64   & 20.10     \\
ParsingSum-HT ($\alpha$=1)       & 36.77   & 12.93   & 20.35     \\
ParsingSum-HT ($\alpha$=2)       & 36.27   & 12.27   & 19.93     \\
ParsingSum-HT ($\alpha$=3)         & \textbf{37.34}  & \textbf{13.00} & \textbf{20.42} \\ \hline
FT                      & 42.98   & 14.48   & 20.06  \\
ParsingSum-FT ($\alpha$=1)       & 44.03   & 15.00   & 20.21    \\
ParsingSum-FT ($\alpha$=2)      & \textbf{44.32}   & \textbf{15.23}   & \textbf{20.72}    \\
ParsingSum-FT ($\alpha$=3)         & 44.02   & 15.20 & 20.31     \\ \hline
\end{tabular}
\end{table}

\subsubsection{The Analysis of the Fusion Weights in Linguistic-guided Attention}

In ParsingSum, the trade-off factor $\alpha$ controls the intensity of attention from linguistic perspective and the multi-head attention. 
To analyze its importance and influence, we conducted experiments by setting $\alpha$ to 0, 1, 2 and 3 ($\alpha$ = 0 denotes the naive Transformer model without linguistic-guided attention) on 
the two versions of ParsingSum. The results are shown in Table \ref{tab:parsing_weight_HT_b13000} where the ROUGE-1, ROUGE-2 and ROUGE-L scores are reported.
Noted that the results show that the ParsingSum model performs differently with different $\alpha$.
Generally, there is an increasing trend with the increment of $\alpha$. This rising trend further prove assigning a relative larger $\alpha$ in a suitable range is able to improve the performance of summarization models. ParsingSum-HT achieves the best result when $\alpha$ =3, and ParsingSum-HT achieves the best performance when when $\alpha$ =2.

\subsubsection{The Analysis of Batch Size}

\begin{figure*}[t]
\centering
\subfigure{
\begin{minipage}[t]{0.33\linewidth}
\centering
\includegraphics[width=1\textwidth]{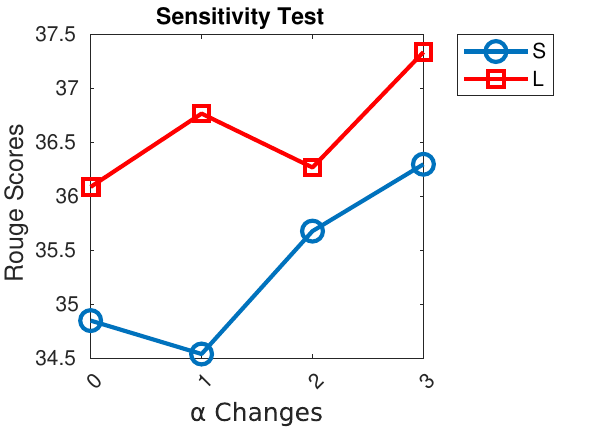}
(a)
\end{minipage}%
    }%
\subfigure{
\begin{minipage}[t]{0.33\linewidth}
\centering
\includegraphics[width=1\textwidth]{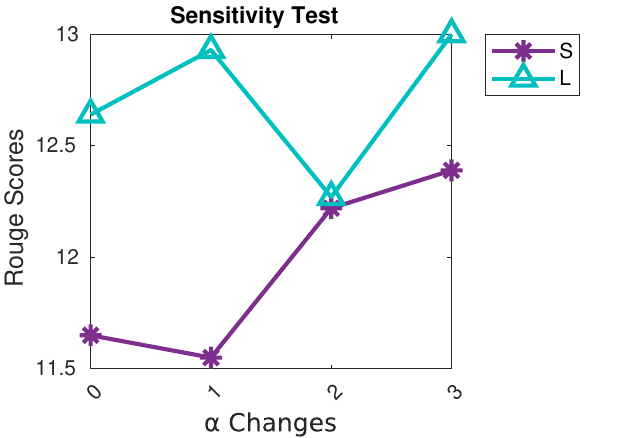}
(b)
\end{minipage}%
}%
\subfigure{
\begin{minipage}[t]{0.33\linewidth}
\centering
\includegraphics[width=1.0\textwidth]{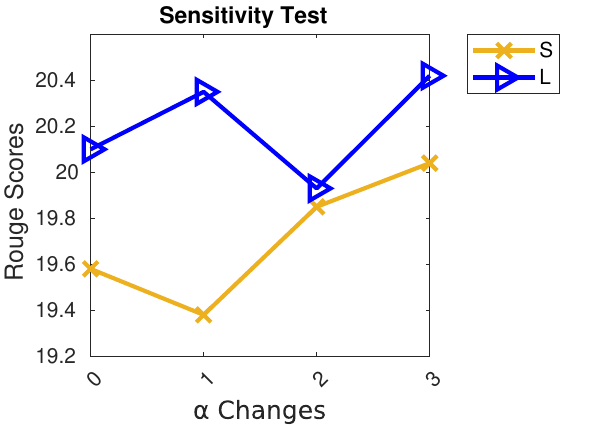}
(c)
\end{minipage}%
}%
\centering
\caption{ParsingSum-HT performances on small (``S'') and large batch size setting (``L''). }
\label{fig:small_batchsize}
\end{figure*}

The batch size is considered to affect the training and testing process of a deep learning model, thus further affects the model's performance. To validate this empirically, we adjusted the batch size of HT model and ParsingSum-HT to a smaller batch size 4,500 (the large batch size is 13,000) and tested them with different trade-off factor $\alpha$. The results are shown in Figure \ref{fig:small_batchsize}. As shown, smaller batch size reduces the model performance on all the evaluation metrics. Interestingly, in the small batch size setting, the ROUGE scores are steadily increasing with $\alpha$ changes from 1 to 3; while trained with large batch size, the increasing trend are retained but the ROUGE scores are jittering when $\alpha$ equals to two. It indicates different batch sizes have different sensitivities towards the change of $\alpha$.

\subsection{Analysis}

\subsubsection{Analysis of the Fusion Methods for Parsing information}

How to integrate the parsing information into the Transformer-based model is an important task at the beginning of the work. 
In addition to the fusion method introduced in Section \ref{Linguistic Guided Encoding Layer}, we attempted several other fusion methods under small batch size setting of the ParsingSum-HT model.

\noindent \textbf{Direct fusion.} In this method,  denoted as ParsingSum-HT (P0.25), we weighted the dependency parsing matrix and added it directly to  multi-head attention :

\begin{equation} \small
LGAtt_{ij}=0.25\mathbb{P}_{i j}+ att_{ij}
\end{equation}

\noindent \textbf{Gaussian-based fusion}. We adopted the idea from \cite{wei2020leveraging} and applied  Gaussian weights to the product of the dependency information matrix and the multi-head attention. In this experiment, the Gaussian weights are set to 0.25 and 8 and denoted as ParsingSum-HT (G0.25) and ParsingSum-HT (G8) respectively:

\begin{equation} \small
LGAtt_{ij}=\frac{(1- att_{ij}\mathbb{P}_{i j})^{2}}{0.25} +att_{ij}
\end{equation}

\begin{equation} \small
LGAtt_{ij}=\frac{(1- att_{ij}\mathbb{P}_{i j})^{2}}{8} +att_{ij}
\end{equation}

Table \ref{tab:different_method_adding_parsing} presents the performances of the mentioned fusion methods. We see our method ParsingSum-HT with $\alpha$=3  receives the best results for ROUGE-1, ROUGE-2 and ROUGE-L scores. 
The potential reason is through direct weighted (including Gaussian weighted) sum of dependency parsing matrix and multi-head attention, the scale of the original multi-head attention has been lost, which leading to pose the dependency parsing matrix in a dominate position. In this case, the normal gradient back propagation process has been disturbed. This experiment indicates that a direct summation of the weighed dependency parsing matrix and multi-head attention may damage the original attention. Thus, a ``soft'' fusion (when $\alpha$ is adopted as mentioned in Section \ref{Linguistic Guided Encoding Layer}) of these two attentions can achieve the best results.

\begin{table} \small
\caption{Performance of ParsingSum-HT via Different Fusion Methods.}
\centering
\label{tab:different_method_adding_parsing}
\setlength{\tabcolsep}{2.5mm}
\begin{tabular}{c|ccc}
\hline
\multirow{2}{*}{Models} & \multicolumn{3}{|c}{ROUGE-F}  \\
                        & 1       & 2       & L         \\ \hline
ParsingSum-HT (P0.25)       & 19.50    & 3.40     & 12.59    \\
ParsingSum-HT (G0.25)       & 16.84   & 1.92    & 11.36     \\
ParsingSum-HT (G8)          & 20.18   & 3.55    & 13.00       \\
ParsingSum-HT ($\alpha$=3)         & \textbf{36.30}    & \textbf{12.39}   & \textbf{20.04}     \\ \hline
\end{tabular}
\end{table}

\begin{figure*}[htp]
\centering
\includegraphics[width=0.8\textwidth]{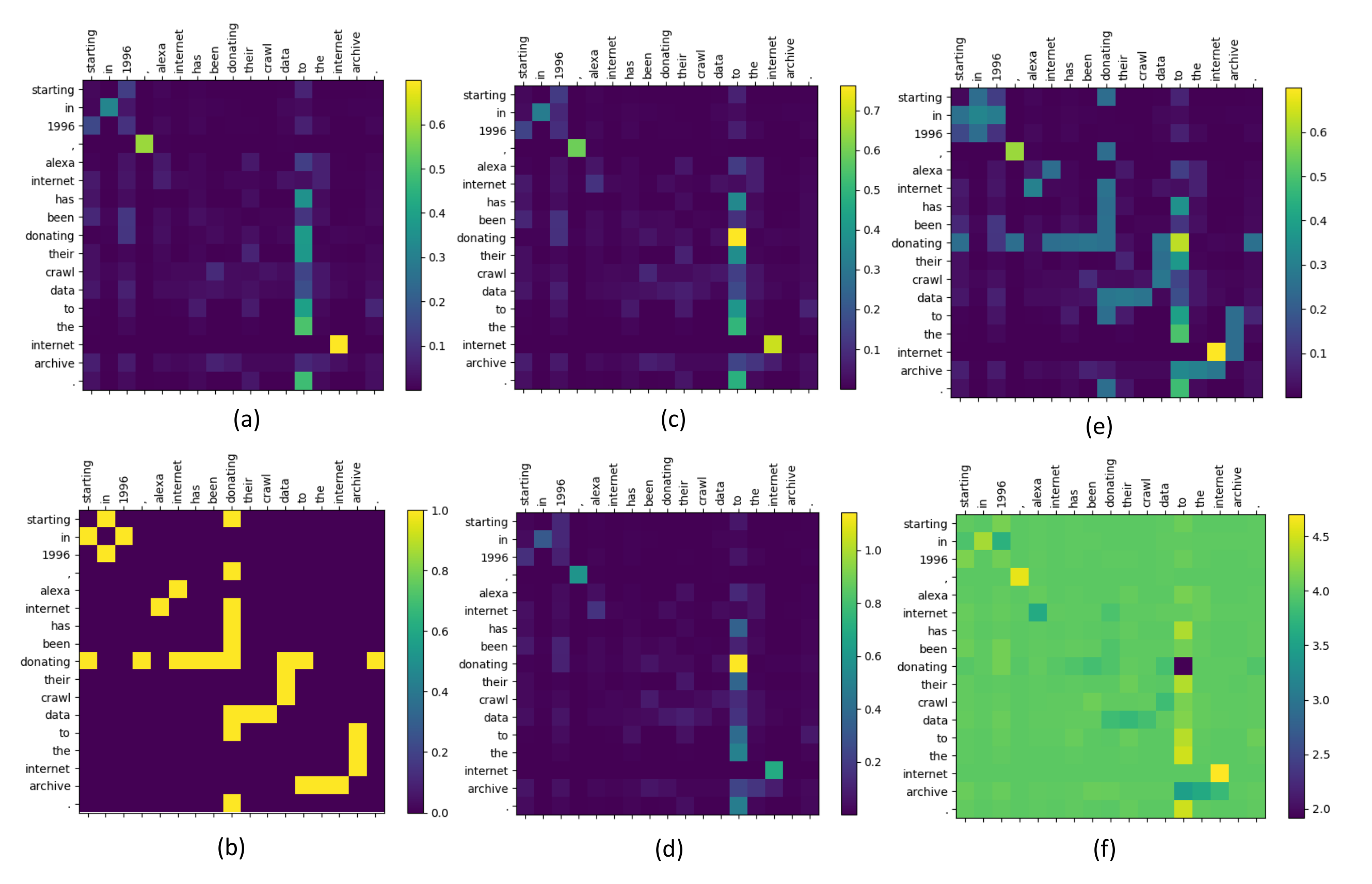}
\caption{Attention Maps. (a) HT model; (b) Heatmap of dependency parsing matrix;  (c) ParsingSum-HT ($\alpha$=1); (d) ParsingSum-HT ($\alpha$=2); (e) ParsingSum-HT (P0.25); (f)  ParsingSum-HT (G0.25). }
\label{fig:heatmap}
\end{figure*}

We further visualized the attention maps for different fusion methods as shown in Figure \ref{fig:heatmap}. 
Figure 4(a) represents the heapmap of multi-head attention within HT model and Figure 4(b) shows the heatmap of our  dependency parsing matrix. Figure 4(c) to 4(f) illustrate the attention maps of different fusion methods. More specifically, subfigure (c) and (d) are the ``soft'' augmentation of multi-head attention with the linguistic-guided knowledge; subfigure (e) and (f) are the ``hard'' combination of multi-head attention with dependency matrices.

\section{Conclusion}

In this paper, we explored the importance of dependency parsing in news document summarization and propose a generic framework to leverage dependency relations among documents for abstractive summarization performance improvement. The experiments show that the proposed model outperforms several strong baselines. It confirms that utilizing dependency relations of the source documents is beneficial to guide the final summaries generation. We also conduct the sensitivity test for choosing the best parsing weights and batch size. Analysis of different ways to incorporate parsing information is further performed to indicate that a ``soft'' method can receive the best results compared to direct weighted sum of dependency parsing matrix and multi-head attention. 
% In the future we will apply our linguistic-guided attention mechanism to other Transformer-based models and examine the influences of other linguistic knowledge on the MDS task. 
\bibliographystyle{splncs04}
\bibliography{emnlp2021}




% LNCS Homepage, \url{http://www.springer.com/lncs}. Last accessed 4
% \end{thebibliography}
\end{document}